\pdfoutput=1

\documentclass[11pt]{article}

\usepackage[]{acl}

\usepackage{times}
\usepackage{latexsym}

\usepackage{array}
\newcolumntype{H}{>{\setbox0=\hbox\bgroup}c<{\egroup}@{}}

\usepackage[T1]{fontenc}

\usepackage{tipa}
\usepackage{hyperref}

\usepackage{microtype}

\newcommand{\tabref}[2][]{Table#1~\ref{tab:#2}}
\newcommand{\figref}[2][]{Figure#1~\ref{fig:#2}}

\newcommand{\equalsign}{\footnotemark[1]\hspace{0.1cm}}

\usepackage{amsmath}
\usepackage{multirow}
\usepackage{bbm}
\usepackage{graphicx}
\usepackage{amssymb}
\usepackage{booktabs}
\usepackage{adjustbox}
\usepackage{xspace}
\usepackage{algpseudocode}
\usepackage{algorithm}
\usepackage{times}
\usepackage{latexsym}
\usepackage{graphicx}
\usepackage{caption}
\usepackage{subcaption}
\usepackage{color, colortbl}
\usepackage{tablefootnote}
\usepackage{longtable}

\newcommand{\STAB}[1]{\begin{tabular}{@{}c@{}}#1\end{tabular}}

\title{One Country, 700+ Languages: NLP Challenges for Underrepresented Languages and Dialects in Indonesia}

\author{Alham Fikri Aji$^1$\thanks{\hspace{0.2cm}These authors contributed equally.}\hspace{0.15cm}\thanks{\hspace{0.2cm}Work done prior to joining Amazon.}\hspace{0.1cm}, Genta Indra Winata$^2$\equalsign, Fajri Koto$^3$\equalsign, Samuel Cahyawijaya$^4$\equalsign,\\ \bf Ade Romadhony$^{5,7}$\equalsign, Rahmad Mahendra$^{6,7}$\equalsign, Kemal Kurniawan$^{3,7}$, David Moeljadi$^8$,\\ \bf Radityo Eko Prasojo$^9$, Timothy Baldwin$^{3,10}$, Jey Han Lau$^3$, Sebastian Ruder$^{11}$\\
$^1$Amazon \quad $^2$Bloomberg \quad $^3$The University of Melbourne \quad $^4$HKUST \\ \quad $^5$Telkom University \quad $^6$Universitas Indonesia \quad $^{7}$INACL \\ \quad $^8$Kanda University of International Studies \quad $^9$Kata.ai  \quad $^{10}$MBZUAI  \quad $^{11}$Google Research \\
\texttt{\scriptsize afaji@amazon.co.uk, gwinata@bloomberg.net, fajri.phd@gmail.com, rahmad.mahendra@cs.ui.ac.id}}

\begin{document}
\maketitle
\begin{abstract}
NLP research is impeded by a lack of resources and awareness of the challenges presented by underrepresented languages and dialects. Focusing on the languages spoken in Indonesia, the second most linguistically diverse and the fourth most populous nation of the world, we provide an overview of the current state of NLP research for Indonesia's 700+ languages. We highlight challenges in Indonesian NLP and how these affect the performance of current NLP systems. Finally, we provide general recommendations to help develop NLP technology not only for languages of Indonesia but also other underrepresented languages.
\end{abstract}


\section{Introduction}

\begin{figure}[!ht]
	\centering
	\begin{subfigure}{0.99\linewidth}
		\centering
		\includegraphics[width=\linewidth]{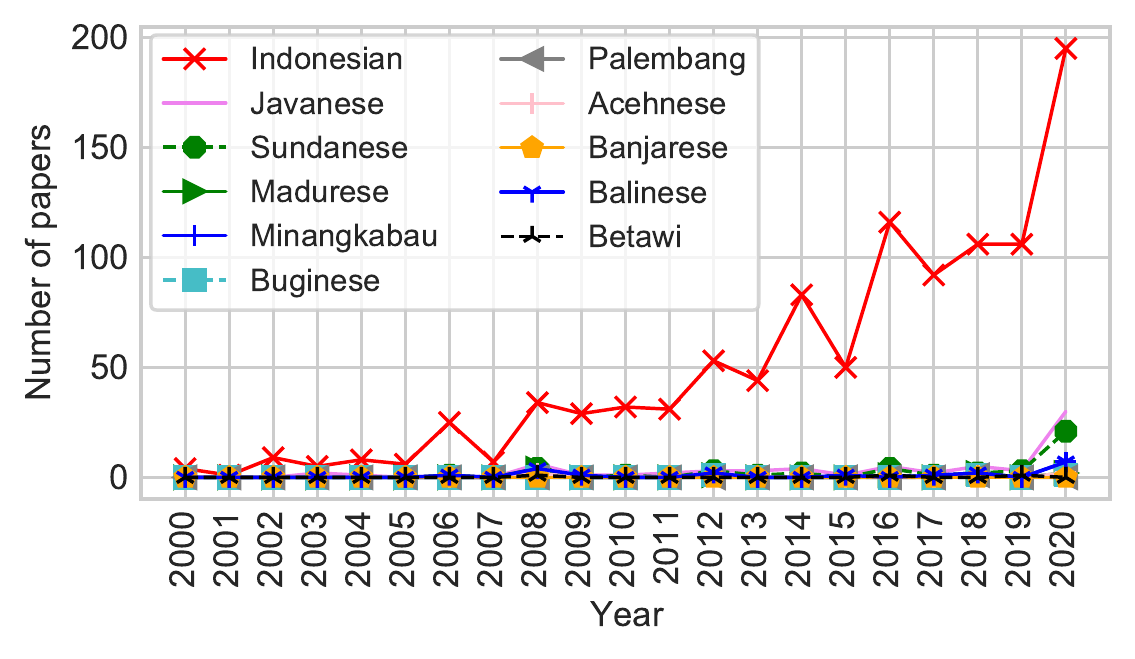}   
	\end{subfigure}
	\begin{subfigure}{0.99\linewidth}
		\centering
		\includegraphics[width=\linewidth]{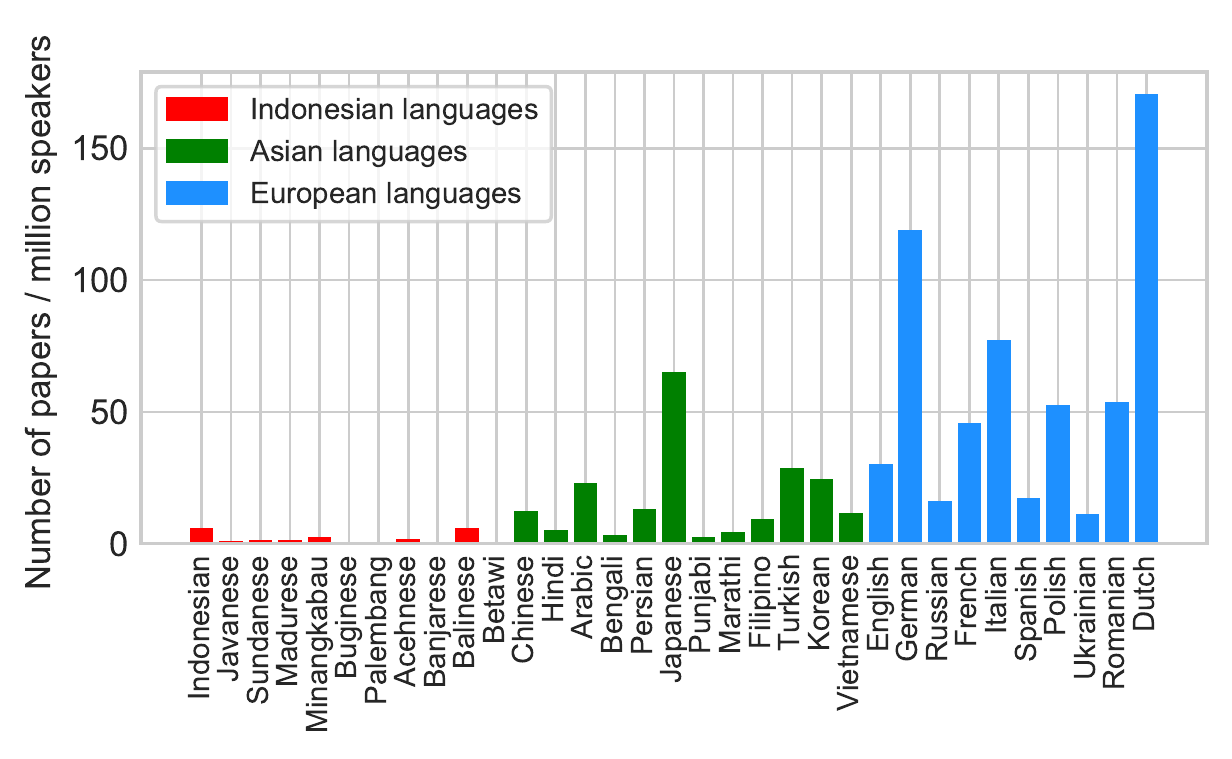} 
	\end{subfigure}
	\caption{Following \citet{Joshi2020}, we compile ACL Anthology to count the distribution of published works that mention languages spoken in Indonesia. \textbf{Top:} Distribution of papers in 20 years. \textbf{Bottom:} Number of papers per a million speakers. We compare languages spoken in Europe, Asia, and Indonesia.
	}
	\label{fig:stats}
	\vspace{-0.5cm}
\end{figure}

Research in natural language processing (NLP) has traditionally focused on developing models for English and a small set of other languages with 
large amounts of data (see~Figure \ref{fig:stats}, bottom right). 
While the lack of data is generally cited as the key reason for the lack of progress in NLP for underrepresented languages \cite{Hu2020xtreme,Joshi2020}, we argue that another factor relates to the diversity and the lack of understanding of the linguistic characteristics of such languages.
Through the lens of the languages spoken in Indonesia, the world's second-most linguistically diverse country, we seek to illustrate 
the challenges
in applying NLP technology to such a diverse pool of languages.

Indonesia is the 4th most populous nation globally, with 273 million people spread over 17,508 islands.
There are more than 700 languages spoken in Indonesia, equal to 10\% of the world's languages, second only to Papua New Guinea~\cite{ethnologue}. However, most of these languages are not well documented in the literature; many are not formally taught, and no established standard exists across speakers \cite{novitasari-etal-2020-cross}. 
Many of them are decreasing in use, as Indonesian (\textit{Bahasa Indonesia}), the national language, is more frequently used as the primary language across the country. 
This process may ultimately result in a monolingual society~\cite{cohn2014local}.

\begin{figure*}[!t]
\centering
\begin{minipage}{.45\textwidth}
  \centering
  \includegraphics[width=.95\linewidth]{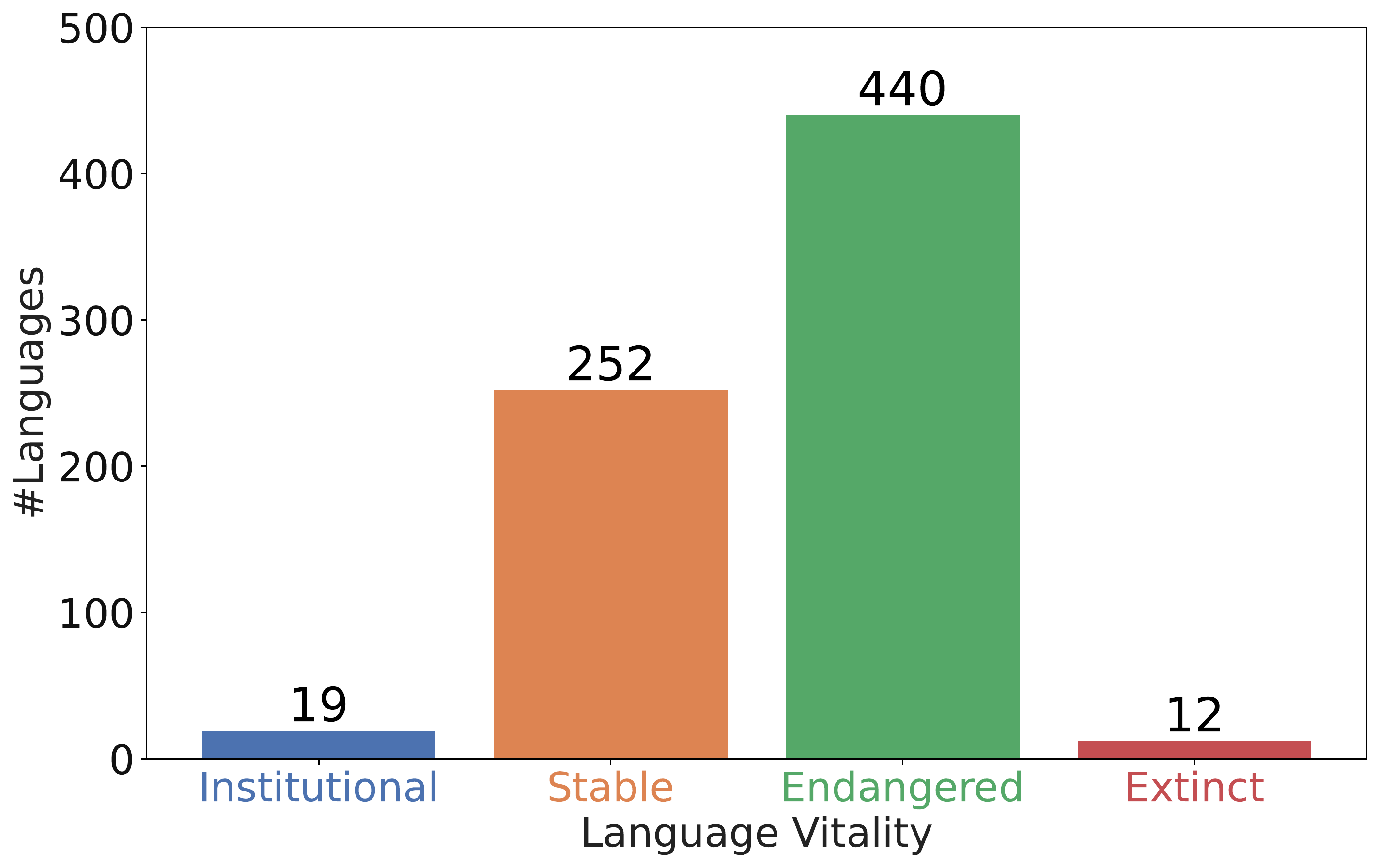}  
\end{minipage}
\begin{minipage}{.45\textwidth}
  \centering
  \includegraphics[width=.95\linewidth]{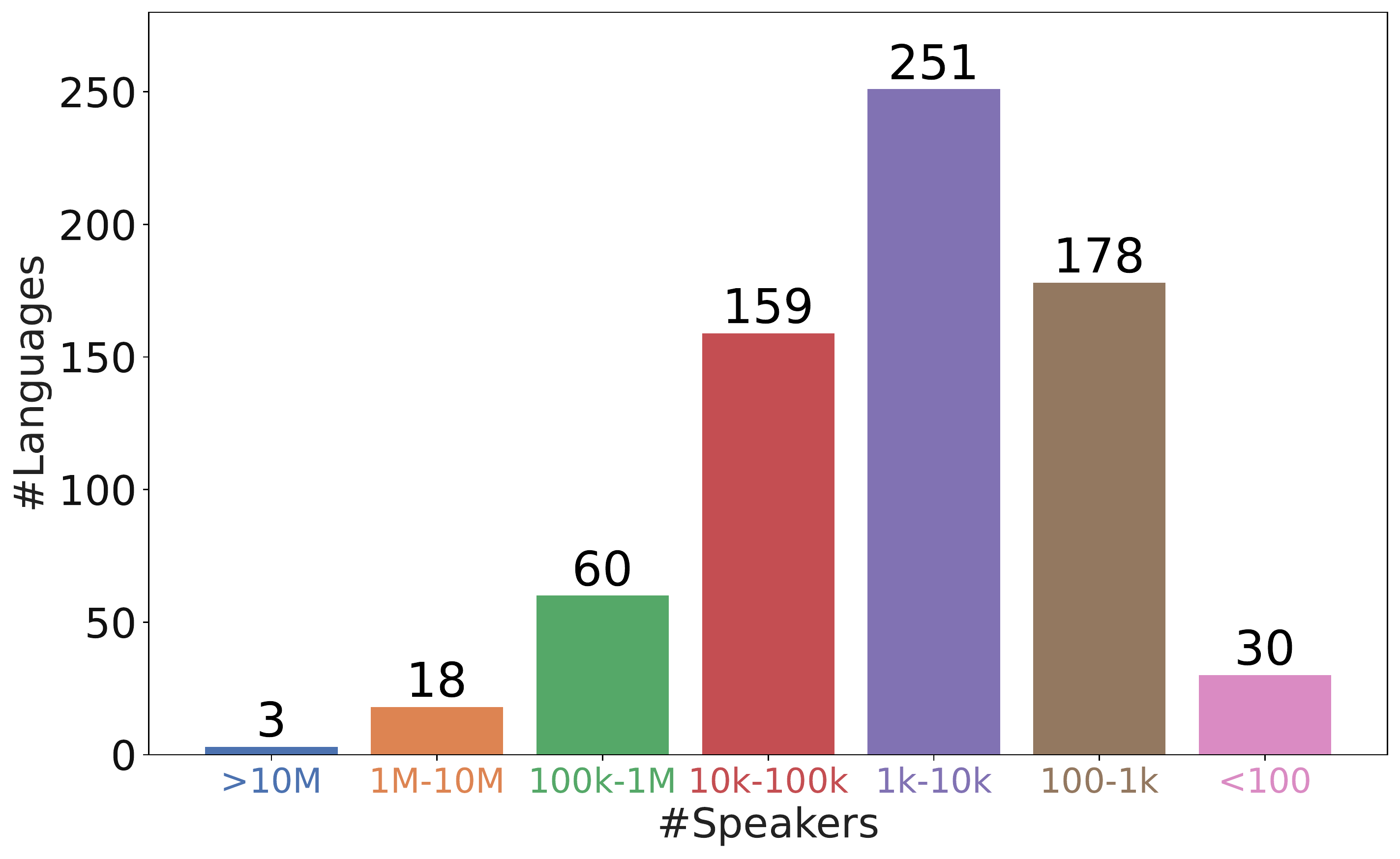}
\end{minipage}
\caption{Distribution of 700+ languages spoken in Indonesia according to Ethnologue~\cite{ethnologue}. \textbf{Left:} Language vitality. \textbf{Right:} Speaker count.}
\label{fig:vitality}
\end{figure*}

Among more than 700 Indonesian local languages, many are threatened. 440 languages are listed as endangered and 12 as extinct according to data from Ethnologue~\cite{ethnologue} illustrated in Figure~\ref{fig:vitality}. \citet{anindyatri} found nearly half of a sample of 98 Indonesian local languages to be endangered while \citet{Esch2022} observed 71 among 151 Indonesian local languages to have less than 100k speakers.



\begin{table}[!t]
    \centering
    \small
    \begin{tabular}{lcr}
    \toprule
        \textbf{Language} & \textbf{ISO} & \textbf{\# Speakers }\\
    \midrule
        Indonesian & ind & 198 M \\
        Javanese & jav & 84 M \\
        Sundanese / Sunda & sun & 34 M \\
        Madurese / Madura & mad & 7 M \\
        Minangkabau & min & 6 M \\
        Buginese & bug & 6 M \\ 
        Betawi & bew  & 5 M \\
        Acehnese / Aceh & ace &  4 M \\ 
        Banjar & bjn &  4 M \\ 
        Balinese & ban  & 3 M \\
        Palembang Malay (Musi) & mus & 3 M \\
    \bottomrule
    \end{tabular}
    \caption{The number of speakers for Indonesian and top-10 most spoken local languages in Indonesia \cite{ethnologue}.}
    \label{tab:population}
\end{table}



Table~\ref{tab:population} lists the names of the 10 most spoken local languages in Indonesia~\cite{ethnologue}. Javanese and Sundanese are at the top with 84M and 34M speakers, respectively,
while Madura, Minangkabau, and Buginese each have around 6M speakers. Despite their large speaker populations, these local languages are poorly represented in the NLP literature.
Compared to Indonesian, the number of research papers mentioning these languages has barely increased over the past 20 years (Figure \ref{fig:stats}, top).
Furthermore, compared to their European counterparts, Indonesian languages are drastically understudied (Figure \ref{fig:stats}, bottom).
This is true even for Indonesian, which has nearly 200M speakers.

Language technology should be accessible to everyone in their native languages \cite{elra}, including Indonesians.
In the context of Indonesia, language technology research offers some benefits.
First, language technology is a potential peacemaker tools in a multi-ethnic country, helping Indonesians understand each other better and avoid the ethnic conflicts of the past~\cite{bertrand2004nationalism}. On a
larger
scale, language technology promotes language use~\cite{elra} and helps language preservation. Despite these benefits, following \citet{bird-2020-decolonising}, we recommend a careful assessment of individual usage scenarios of language technology, so they are implemented for the good of the local population.


For language technology to be useful in the Indonesian context, it
additionally has to account for the
dialects of local languages. Language dialects in Indonesia are influenced by the geographical location and regional culture of their speakers~\cite{vander2015dichotomy} and thus often differ substantially in morphology and vocabulary, posing challenges for NLP systems. In this paper, we provide an overview of the current state of NLP for Indonesian and Indonesia's hundreds of languages. We then discuss the challenges presented by those languages and demonstrate how they affect state-of-the-art systems in NLP. We finally provide recommendations for developing better NLP technology not only for the languages in Indonesia but also for other underrepresented languages.


\section{Background and Related Work \label{sec:background}}

\subsection{History and Taxonomy}



\begin{figure}[!th] 
	\centering
	\includegraphics[width=\linewidth]{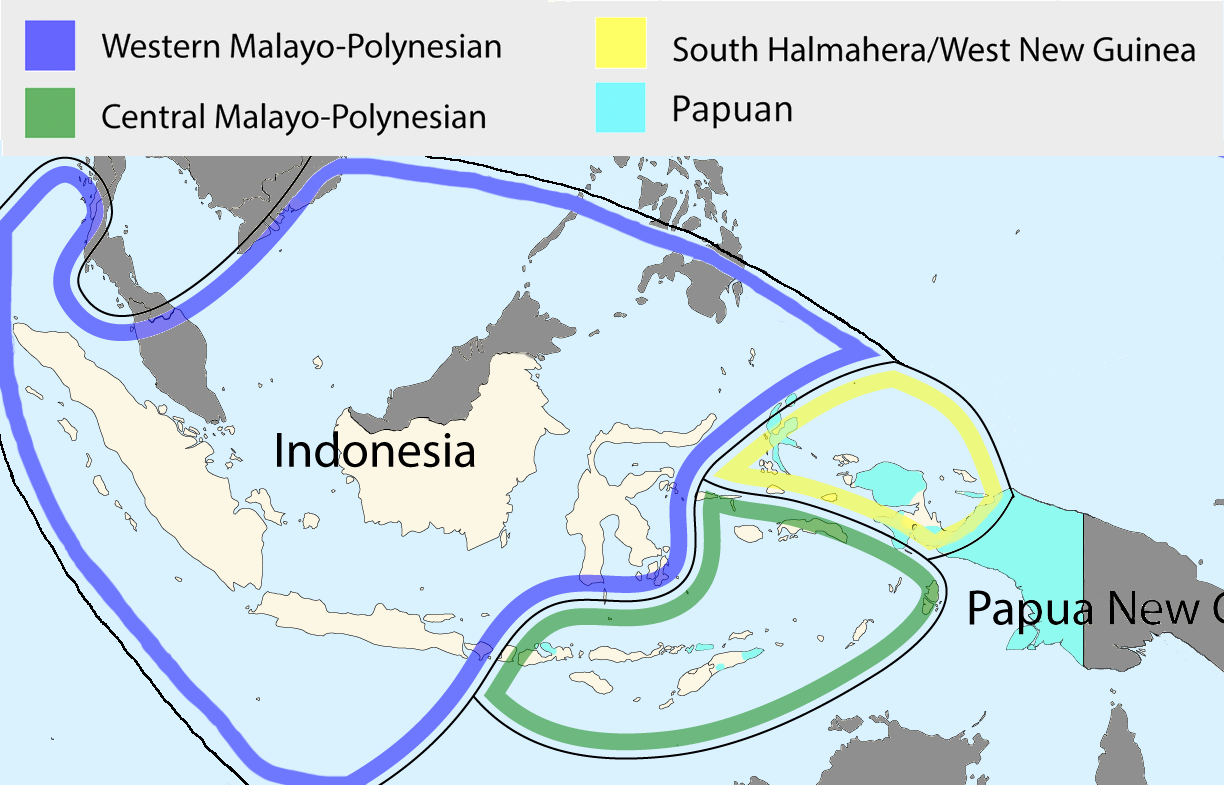}   
	\caption{Map of Austronesian and Papuan languages in Indonesia.}
	\label{fig:map}
\end{figure}








Indonesia is one of the richest countries globally in terms of linguistic diversity. 
More than 400 of its languages belong to the Austronesian language family, while 
the others are Papuan languages spoken in the eastern part of the country. 
As shown in \figref{map}, the Austronesian languages in Indonesia belong to three main groups: Western-Malayo-Polynesian (WMP), Central-Malayo-Polynesian (CMP), and South-Halmahera-West-New-Guinea (SHWNG)~\cite{blust1980austronesian}.
WMP languages are Malay, Indonesian, Javanese, Sundanese, Balinese, and Minangkabau, among others. All languages mentioned in Table~\ref{tab:population} are in this group. 
Languages belonging to CMP are languages of the Lesser Sunda Islands from East Sumbawa (with Bimanese) onwards to the east, and languages of the central and southern Moluccas (including the Aru Islands and the Sula Archipelago). 
The SHWNG group consists of languages of Halmahera and Cenderawasih Bay, and further-flung regions such as the Mamberamo River and the Raja Ampat Islands.
Meanwhile, the Papuan languages are mainly spoken in Papua, such as Dani, Asmat, Maybrat, and Sentani. Some Papuan languages are also spoken in Halmahera, Timor, and the Alor Archipelago~\cite{palmer2018languages,ross2005pronouns}.

Most Austronesian linguists and archaeologists agree that the original `homeland' of Austronesian languages must be sought in Taiwan and, prior to Taiwan, in coastal South China~\cite{adelaar:2005,bellwood2011cultures}. 
In the second millennium CE, the Austronesian people moved from Taiwan to the Philippines. From the Philippines, they moved southward to Borneo and Sulawesi. From Borneo, they migrated to Sumatra, the Malay Peninsula, Java, and even to Madagascar. From Sulawesi, they moved southward to the CMP area and eastward to the SHWNG area. From there, they migrated to Oceania and Polynesia, as far as New Zealand, Easter Island, and Hawaii~\cite{gray2000}. The people that lived in insular Southeast Asia, such as in the Philippines and Indonesia, before the arrival of Austronesians were Australo-Melanesians~\cite{bellwood2007prehistory}. Gradual assimilation with Austronesians occurred, although some pre-Austronesian groups still survive, such as Melanesian people in eastern Indonesia~\cite{ross2005pronouns,coupe2020asia}.

At the time of the arrival of the first Europeans, Malay had become the major language (lingua franca) of interethnic communication in Southeast Asia and beyond~\cite{steinhauer:2005,coupe2020asia}. It functioned as the language of trade and the language of Islam because Muslim merchants from India and the Middle East were the first to introduce the religion into the harbor towns of Indonesia. After the arrival of Europeans, Malay was used by the Portuguese and Dutch to spread Catholicism and Protestantism. When the Dutch extended their rule over areas outside Java in the nineteenth century, the importance of Malay increased, and thus, the first standardization of the spelling and grammar occurred in 1901, based on Classical Malay~\cite{abas_indonesian_1987,sneddon2003}. In 1928, the Second National Youth Congress participants proclaimed Malay (henceforth called Indonesian) as the unifying language of Indonesia. During World War II, the Japanese occupying forces forbade all use of Dutch in favor of Indonesian, which from then onward effectively became the new national language. From independence until the present, Indonesian has functioned as the primary language in education, mass media, and government activities. Many local language speakers are increasingly using Indonesian with their children because they believe it will aid them to attain a better education and career~\cite{klamer2018}.

\subsection{Efforts in Multilingual Research}




Recently, pretrained multilingual language models such as mBERT~\cite{devlin-etal-2019-bert}, mBART~\cite{liu2020mbart}, and mT5~\cite{xue-etal-2021-mt5} have been proposed. Their coverage, however, focuses on high-resource languages. Only mBERT and mT5 include Indonesian local languages, i.e., Javanese, Sundanese, and Minangkabau, but with comparatively little pretraining data.

Some multilingual datasets for question answering~\cite[TyDiQA;][]{clark-etal-2020-tydi}, common sense reasoning~\cite[XCOPA;][]{ponti-etal-2020-xcopa}, abstractive summarization~\cite{hasan-etal-2021-xl}, passage ranking~\cite[mMARCO;][]{bonifacio2021mmarco}, cross-lingual visual question answering~\cite[xGQA;][]{Pfeiffer2021xGQACV}, language and vision reasoning~\cite[MaRVL;][]{liu-etal-2021-visually}, paraphrasing~\cite[ParaCotta;][]{aji2021paracotta}, dialogue systems~\cite[XPersona \& BiToD;][]{lin-etal-2021-xpersona,lin2021bitod}, lexical normalization~\cite[MultiLexNorm;][]{van-der-goot-etal-2021-multilexnorm}, and machine translation \cite[FLORES-101;][]{flores1} include Indonesian but most others do not, and very few include Indonesian local languages. An exception is the weakly supervised named entity recognition dataset, WikiAnn~\cite{pan-etal-2017-cross}, which covers several Indonesian local languages, namely Acehnese, Javanese, Minangkabau, and Sundanese. 



Parallel corpora including Indonesian local languages are: (i) CommonCrawl; (ii) Wikipedia parallel corpora like MediaWiki Translations;\footnote{\url{https://mediawiki.org/wiki/Content_translation}} and WikiMatrix~\cite{schwenk-etal-2021-wikimatrix} (iii) the Leipzig corpora~\cite{goldhahn-etal-2012-building}, which include Indonesian, Javanese, Sundanese, Minangkabau, Madurese, Acehnese, Buginese, Banjar, and Balinese; and (iv) JW-300~\cite{agic-vulic-2019-jw300}, which includes dozens of Indonesian local languages, e.g., Batak language groups, Javanese, Dayak language groups, and several languages in Nusa Tenggara. Recent studies, however, have raised concerns regarding the quality of such multilingual corpora for underrepresented languages~\cite{Caswell2022QualityAA}.

\subsection{Progress in Indonesian NLP }

NLP research on Indonesian has occurred across multiple topics, such as POS tagging~\cite{wicaksono2010hmmbp, dinakaramani_pos_tag}, NER~\cite{NER_indra,rachman2017ner,gunawan2018named}, sentiment analysis~\cite{aqsath2011sentiment, lunando2013indonesian, wicaksono2014automatically}, hate speech detection~\cite{alfina2017hate, sutejo2018indonesia}, topic classification~\cite{winata2015handling, kusumaningrum2016}, question answering~\cite{mahendra-etal-2008-extending,fikri2012cdqa}, machine translation~\cite{yulianti_mt, simon2013experiments, hermanto2015recurrent}, keyphrases extraction~\cite{saputra-etal-2018-keyphrases, trisna2020single}, morphological analysis~\cite{pisceldo2008two}, and speech recognition~\cite{lestari2006large,baskoro2008developing,zahra2009building}. 
However, many of these studies either did not release the data or used non-standardized resources with a lack of documentation and open source code, making them extremely difficult to reproduce.

Recently, \citet{wilie-etal-2020-indonlu}, \citet{koto-etal-2020-indolem, koto-etal-2021-indobertweet}, and \citet{cahyawijaya-etal-2021-indonlg} collected Indonesian NLP resources as benchmark data.
Others have also begun to create standardized labeled data for Indonesian NLP, e.g. the works of {\setcitestyle{citesep={,}}\citet{kurniawan2018toward, guntara-etal-2020-benchmarking, koto-etal-2020-liputan6, khairunnisa-etal-2020-towards}}, and \citet{mahendra-etal-2021-indonli}. \setcitestyle{citesep={;}}

On the other hand, there has been very little work on local languages. Several works studied stemming (Sundanese~\cite{suryani2018rule}; Balinese~\cite{subali2019kombinasi}) and POS Tagging \cite[Madurese;][]{dewi2020combination}. \citet{koto-koto-2020-towards} built a Indonesian Minangkabau parallel corpus and also sentiment analysis resources for Minangkabau. Other works developed machine translation systems between Indonesian and local languages, e.g., Sundanese~\cite{suryani2015sundamt},  Buginese~\cite{apriani2016pengaruh}, Dayak Kanayatn~\cite{hasbiansyah2016tuning}, and 
Sambas Malay~\cite{ningtyas2018penggunaan}.


{\setcitestyle{citesep={,}}\citet{tanaya2016dictionary, Tanaya-2018-word}} studied Javanese character segmentation in non-Latin script. \citet{safitri2016spoken} worked on spoken data language identification in Minangkabau, Sundanese, and Javanese, while \citet{azizah2021} developed end-to-end neural text-to-speech models for Indonesian, Sundanese, and Javanese. {\setcitestyle{citesep={,}}\citet{nasution2017generalized, nasution-2021-plan}} proposed an approach for bilingual lexicon induction and evaluated the approach on seven languages, i.e.,  Indonesian, Malay, Minangkabau, Palembang Malay, Banjar, Javanese, and Sundanese.

\citet{cahyawijaya-etal-2021-indonlg} established a machine translation benchmark in Sundanese and Javanese using Bible data. ~\citet{wibowo-etal-2021-indocollex} studied a family of colloquial Indonesian, which is influenced by some local languages via morphological transformation, and \citet{Putri_etal_2021_abusive} worked on abusive language and hate speech detection on Twitter for five local languages, namely Javanese, Sundanese, Madurese, Minangkabau, and Musi.

\section{Challenges for Indonesian NLP}
\subsection{Limited Resources}
\label{sec:limited-res}

\begin{figure}[!th]
    \centering
    \includegraphics[width=0.95\linewidth]{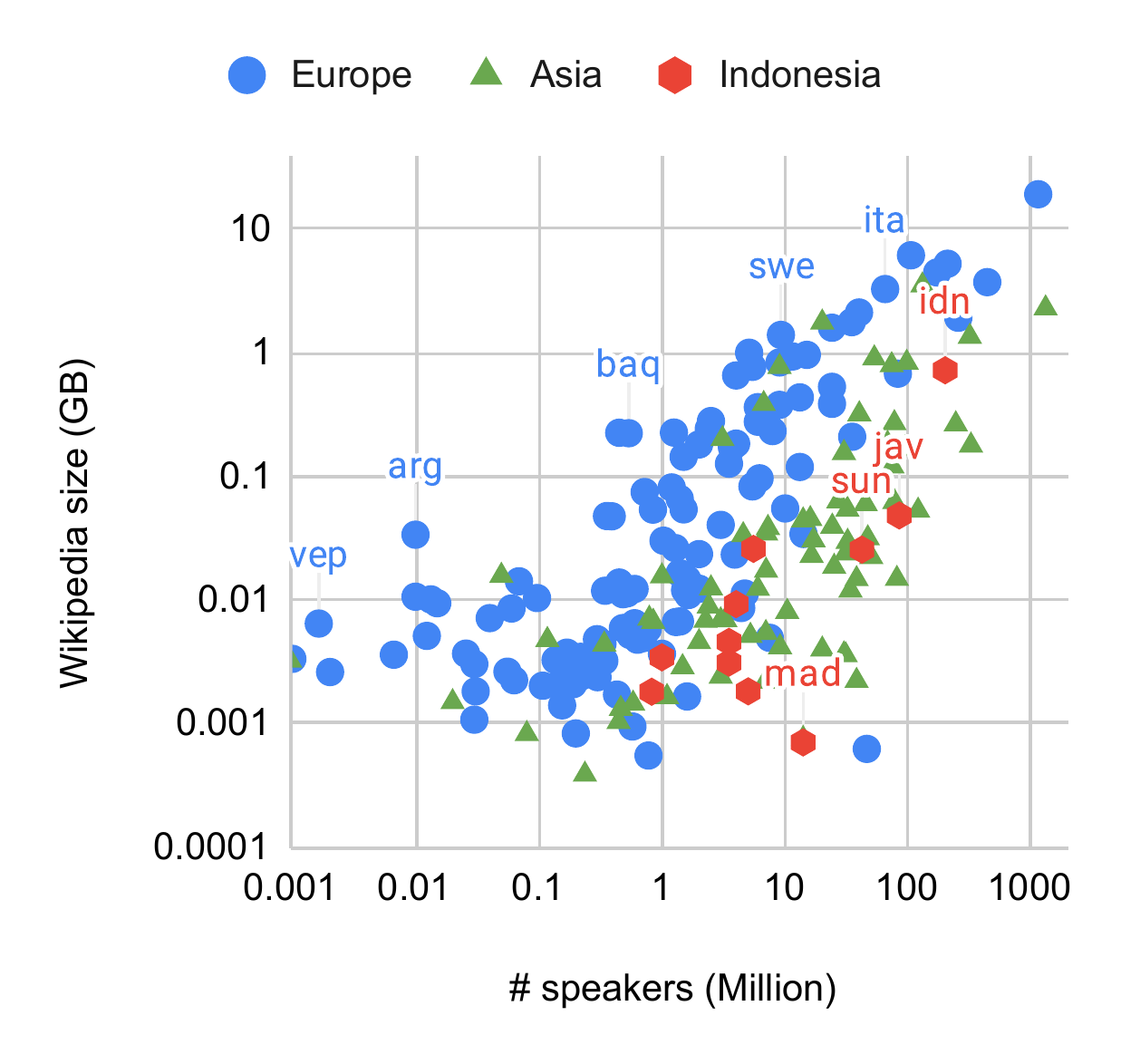}
    \caption{
    Relationship between the number of speakers and the size of data in Wikipedia for languages spoken in Europe, Asia, and Indonesia.
    }
    \label{fig:wiki_vs_speaker}
    \vspace{-0.3cm}
\end{figure}




\paragraph{Monolingual Data}
Unlabeled corpora are crucial for building large language models, such as GPT-2~\cite{radford2019language} or BERT~\cite{devlin-etal-2019-bert}. 
Available unlabeled corpora such as Indo4B \cite{wilie-etal-2020-indonlu}, and Indo4B-Plus \cite{cahyawijaya-etal-2021-indonlg} mainly include data in Indonesian, with the latter containing $\approx$10\% of data in Javanese and Sundanese. 

In comparison, in multilingual corpora such as CC--100 \cite{conneau-etal-2020-unsupervised}, Javanese and Sundanese data accounts for only 0.001\% and 0.002\% of the corpus size respectively while in mC4~\cite{xue-etal-2021-mt5}, there are only 0.6M Javanese and 0.3M Sundanese tokens out of a total of 6.3T tokens.
In addition, we measure data availability in Wikipedia compared to the number of speakers in Figure~\ref{fig:wiki_vs_speaker}.\footnote{The number of speakers is collected from Wikidata~\cite{vrandevcic2014wikidata}, from the \texttt{number of speakers (P1098)} property as of Nov 7th, 2021, while the size is collected from the 20211101 Wikipedia dump.} Much less data is available for the languages spoken in Indonesia, compared to European languages with similar numbers of speakers. For example, Wikipedia contains more than 3 GB of Italian articles but less than 50 MB of Javanese articles, despite both languages having a comparable number of speakers. Similarly, Sundanese has less than 25 MB of articles, whereas languages with comparable numbers of speakers have more than 1.5 GB of articles. Similar trends hold for most other Asian languages. Languages in Africa are even more underrepresented in terms of Wikipedia data (see Appendix \ref{appendix:wiki}).


Beyond the highly spoken local languages, most other Indonesian local languages do not have Wikipedia instances, in contrast to European languages with few speakers. It is very difficult to find alternative sources for high-quality text data for other local languages of Indonesia (such as news websites), as most such sources are written in Indonesian. Resources in long-tail languages are even more scarce due to a very low number of speakers. Moreover, most of the languages in the long tail are mainly used in a spoken context, making text data challenging to obtain. These statistics demonstrate that collecting unlabeled corpora for Indonesian local languages is extremely difficult. This makes it impractical to develop strong pretrained language models for these languages, which have been the foundation for many recent NLP systems.

\begin{table*}[ht!]
    \centering
    \small
    \resizebox{\linewidth}{!}{
        \begin{tabular}{@{}lcccccccc@{}}
        \toprule
            \textbf{English} & \textbf{Mudung Laut} & \textbf{Dusun Teluk} & \textbf{Mersam} & \textbf{Suo Suo} & \textbf{Teluk Kuali} & \textbf{Lubuk Telau} & \textbf{Bunga Tanjung} & \textbf{Pulau Aro} \\
        \midrule
            I/me & sayo & aku & \textipa{awaP} & sayo & kito, \textipa{awaP} & am$^b$o & ambo & ambo \\
            You & kau, kamu & kau & ka$^d$n & kamu & kaan & kamu & \textipa{aN, kau, kayo} & \textipa{baPaN} \\
            he/she  &  \textipa{dioP} & \textipa{dioP, {\textltailn}o} & \textipa{{\textltailn}o} & kau & \textipa{{\textltailn}o} & \textipa{{\textltailn}o} & \textipa{{\textltailn}o} & \textipa{i{\textltailn}o}\\
            if & kalu & jiko, kalu & kalu & bilao & kalu & jiko & \textipa{koP} & kalu \\
            one & satu & \textipa{sekoP} & \textipa{sekoP} &  \textipa{sekoP} & \textipa{ci3P} &  \textipa{sekoP} &  \textipa{sekoP}, so &  \textipa{sekoP} \\
        \bottomrule
        \end{tabular}
    }

    \caption{Lexical variation of Jambi Malay across different villages in Jambi~\citep{anderbeck2008malay}.} 
    \label{tab:jambi-word}
    
~\\
    \resizebox{\linewidth}{!}{
        \begin{tabular}{@{}llllll@{}}
        \toprule
            \multirow{2}{*}{\textbf{English}} & \multirow{2}{*}{\textbf{Context}} & \multicolumn{3}{c}{\textbf{Ngoko}} & \textbf{Krama} \\ \cmidrule(l{2pt}r{2pt}){3-5} 
            \cmidrule(l{2pt}r{2pt}){6-6}
             & & Western & Central & Eastern & Eastern \\
        \midrule
            I/me & I like to eat fried rice. & inyong, enyong & aku & aku & kulo \\
            You & Where will you go? & rika, kowe, ko & kowe, siro, sampeyan & koen, awakmu, sampeyan & panjenengan \\
            How & How do I read this? & priwe & piye & yo'opo & pripun \\  
            Why & Why is this door broken? & ngapa & ngopo & opo'o & punapa \\
            Will & Where will you go? & arep & arep & kate, ate & badhe \\
            Not/no & The calculation is not correct. & ora & ora & gak & mboten \\

            \bottomrule
        \end{tabular}
    }
    \caption{Lexical variations of Javanese dialects and styles across different regions of the Java island. Native speakers are asked to translate the words, given the context.}
    \label{tab:jv-word}
\end{table*}

\paragraph{Labeled Data}

Most work on Indonesian NLP (see \textsection \ref{sec:background}) has not publicly released the data or models, limiting reproducibility. Although recent Indonesian NLP benchmarks are addressing this issue, they mostly focus on the Indonesian language (see Appendix~\ref{appendix:resource}).
Some widely spoken local languages such as Javanese, Sundanese, or Minangkabau have extremely small labeled datasets compared to Indonesian, while others have barely any. 

The lack of such datasets makes NLP development for the local languages difficult. 
However, constructing new labeled datasets is still challenging due to: (1) the lack of speakers of some languages; (2) the vast continuum of dialectical variation (see \textsection \ref{sec:dialect_style}); and (3) the absence of writing standard in most local languages (see \textsection \ref{sec:non-standard}).


\subsection{Language Diversity}
The diversity of Indonesian languages is not only reflected in the large number of local languages but also the large number of dialects of these languages (\textsection \ref{sec:dialect_style}). Speakers of local languages also often mix languages in conversation, which makes colloquial Indonesian more diverse (\textsection \ref{sec:code-mixing}). In addition, some local languages are more commonly used in conversational contexts, so they do not have consistent writing forms in written media (\textsection \ref{sec:non-standard}).

\subsubsection{Regional Dialects and Style Differences} \label{sec:dialect_style}

Indonesian local languages often have multiple dialects, depending on the geographical location. Local languages of Indonesian spoken in different locations might be different (have some lexical variation) to one another, despite still being categorized as the same language~\cite{fauzi2018dialect}. For example, 
\citet{anderbeck2008malay} showed that villages across the Jambi province use different dialects of Jambi Malay. Similarly, \citet{kartikasari2018study} showed that Javanese between different cities in central and eastern Java could have more than 50\% lexical variation, while~\citet{purwaningsih2017geografi} showed that Javanese in different districts in the Lamongan has up to 13\% lexical variation. Similar studies have been conducted on other languages, such as Balinese~\cite{maharani2018variasi} and Sasak~\cite{sarwadi2019lexical}.

Moreover, Indonesian and its local languages have multiple styles, even within the same dialect. One factor that affects style is the level of politeness and formality---similar to Japanese and other Asian languages~\cite{bond2016introduction}. More polite language is used when speaking to a person with a higher social position, especially to elders, seniors, and sometimes strangers. Different politeness levels manifest in the use of different honorifics and even different lexical terms.

To illustrate the distinctions between regional dialects and styles, we highlight common words and utterances across dialects and styles in Jambi Malay and Javanese in Tables \ref{tab:jambi-word} and \ref{tab:jv-word} respectively. For Jambi Malay, we sample the result from a prior work~\citep{anderbeck2008malay}. 
For Javanese, we ask native speakers to translate basic words into three regional dialects: Western, Central, and Eastern Javanese, and two different styles: \textit{Ngoko} (standard, daily-use Javanese) and \textit{Krama} (polite Javanese, used to communicate to elders and those with higher social status). However, since contemporary \textit{Krama} Javanese is not very different among regions, we only consider \textit{Krama} from the Eastern speakers' perspective. 

Jambi Malay has many dialects across villages. As shown in Table \ref{tab:jambi-word}, many common words are spoken differently across dialects and styles. Similarly, Javanese is also different across regions. Not every Javanese speaker understands \textit{Krama}, since its usage is very limited. Moreover, the number of Javanese speakers who can use \textit{Krama} is declining~\cite{cohn2014local}.\footnote{\textit{Krama} is used to speak formally (e.g., with older or respected people). However, people prefer to use Indonesian more in a formal situation. People who move from sub-urban areas to bigger cities tend to continue to use \textit{Ngoko} and thus also pass \textit{Ngoko} on to their children.} Examples from other languages are shown in Appendix~\ref{appendix:dialect}.


\begin{table}[!t]
\small
\centering

  \resizebox{0.49\textwidth}{!}{
      \begin{tabular}{@{}c@{ ~ }c@{ ~ }c@{ ~ }c@{ ~ }c@{ ~ }c@{ ~ }c@{}}
      \toprule
         & \multicolumn{5}{c}{\textbf{Model}} \\ \midrule
        \multirow{2}{*}{\textbf{Style}} &\multirow{2}{*}{\textbf{Region}} & \multicolumn{2}{c}{\textbf{langid.py}} & \multicolumn{2}{c}{\textbf{FastText}} & \textbf{CLD3} \\ \cmidrule(l{2pt}r{2pt}){3-4} \cmidrule(l{2pt}r{2pt}){5-6} \cmidrule(l{2pt}r{2pt}){7-7} 
         & & Top-1 & Top-3 & Top-1 & Top-3 & Top-1 \\
         \midrule
        \textit{Ngoko} & Western & 0.241 & 0.621 & 0.069 & 0.379 & 0.759 \\
        \rowcolor{lime} \textit{Ngoko} & Central & 0.345 & 0.690 & 0.379 & 0.724 & 0.828 \\
        \textit{Ngoko} & Eastern & 0.276 & 0.552 & 0.103 & 0.379 & 0.552 \\
        \rowcolor{lime} \textit{Krama} & Eastern & 0.345 & 0.759 & 0.379 & 0.586 & 0.897 \\
        \bottomrule
      \end{tabular} 
  }
\caption{Language identification accuracy based on different Javanese dialects and styles. Systems do not perform equally well across dialects and styles.}
  \label{tab:language-iden-jv}
\end{table}

\subsubsection*{Case Study in Javanese}

Dialectical and style differences pose a challenge to NLP systems. To explore the extent of this challenge, we conduct an experiment to test the robustness of NLP systems to variations in Javanese dialects. We ask native speakers\footnote{Our annotators are based in Banyumas, Jogjakarta, and Jember for Western, Central, and Eastern Javanese respectively. Using dialects from different cities might yield different results.} to translate 29 simple sentences into Javanese according to the specified dialect and style. We then evaluate several 
language identification systems on those instances. Language identification is a core part of multilingual NLP 
and a necessary step for collecting textual data in a language. Despite its importance, it is an open research area, particularly for underrepresented languages \cite{Hughes+:2006,Caswell2022QualityAA}.

We compare langid.py~\cite{lui2012langid}, FastText~\cite{joulin2017bag}, and CLD3.\footnote{\url{https://github.com/google/cld3}} The results can be seen in Table~\ref{tab:language-iden-jv}. In general, the language identification systems are more accurate in detecting Javanese texts in the \textit{Ngoko}-Central dialect, or \textit{Krama}, since the systems were trained on Javanese Wikipedia data, which is written in either the \textit{Ngoko}-Central or \textit{Krama} dialects and styles. If an NLP system can only detect certain dialects, then this information should be conveyed explicitly. Problems arise if we assume that the model works equally well across dialects. For example, in the case of language identification, if we use the model to collect datasets automatically, then Javanese datasets with poor-performing dialects will be underrepresented in the data.

\begin{table}[!t] 
    \centering
    \small
    \resizebox{0.49\textwidth}{!}{
        \begin{tabular}{@{}p{0.44\linewidth}  p{0.47\linewidth}@{}}
            \toprule
             \textbf{Colloquial Indonesian} & \textbf{Translation} \\
             \midrule
             Ada yang \textbf{\textcolor{orange}{nge}\textcolor{red}{tag}} foto \textbf{\textcolor{blue}{lawas}} di FB & Someone is tagging old photos in FB \\
             \textbf{\textcolor{red}{Quote}}nya Andrew Ng ini relevan \textbf{\textcolor{orange}{banget}} & This Andrew Ng quote is very relevant \\ 
             \textcolor{teal}{Bilo} kita pergi main lagi? & When will we go play again? \\ 
             Ini \textbf{\textcolor{olive}{teh}} aksara jawa kenapa susah \textbf{\textcolor{orange}{banget}}? & Why is this Javanese script very difficult? \\
             \bottomrule
        \end{tabular}
    }
    \caption{Colloquial Indonesian code-mixing examples from social media. Color code: \textcolor{red}{English}, \textcolor{orange}{Betawinese}, \textcolor{blue}{Javanese}, \textcolor{teal}{Minangkabau}, \textcolor{olive}{Sundanese}, Indonesian.}
    \label{tab:codemix}
\end{table}

\subsubsection{Code-Mixing} \label{sec:code-mixing}
Code-mixing is an occurrence where a person speaks alternately in two or more languages in a conversation~\cite{sitaram2019survey,winata2018code,winata2019code,dogruoz-etal-2021-survey}. This phenomenon is common in Indonesian conversations~\cite{barik-etal-2019-normalization, johanes2020structuring, wibowo-etal-2021-indocollex}. In a conversational context, people sometimes mix their local languages with standard Indonesian, resulting in colloquial Indonesian~\cite{siregar2014code}. This colloquial-style Indonesian is used daily in speech and conversation and is common on social media~\cite{sutrisno2019beyond}. Some frequently used code-mixed words (especially on social media) are even intelligible to people that do not speak the original local languages. Interestingly, code-mixing can also occur in border areas where people are exposed to multiple languages, therefore mixing them together. For example, people in Jember (a regency district in East Java) combine Javanese and Madurese in their daily conversation~\cite{haryono2012perubahan}.

Indonesian code-mixing not only occurs at the word level but also 
at the morpheme level~\cite{winata2021multilingual}. For example, \textit{quotenya} (``his/her quote'', see Table~\ref{tab:codemix}) combines the English word \textit{quote} and the Indonesian suffix \textit{-nya}, which denotes possession; similarly, \textit{ngetag} combines the Betawinese prefix \textit{nge-} and the English word \textit{tag}. More examples can be found in Table~\ref{tab:codemix}.

\subsection{Orthography Variation} \label{sec:non-standard}

Many Indonesian local languages are mainly used in spoken settings and have no established standard orthography system. Some local languages do originally have their own archaic writing systems that derive from the Jawi alphabet or Kawi script, and even though standard transliteration into the Roman alphabet exists for some (e.g., Javanese and Sundanese), they are not widely known and practiced~\cite{Diksi5265}. Hence, some words have multiple romanized orthographies that are mutually intelligible, as they are pronounced the same. Some examples can be seen in Table~\ref{tab:written-form}. Such a variety of written forms is common in local languages in Indonesia. This variation leads to a significantly larger vocabulary size, especially for NLP systems that use word-based representations, and presents a challenge to constrain the representations for different spellings of the same word to be similar.

\begin{table}[!t] 
    \centering
    \setlength\tabcolsep{1.5pt}
    \resizebox{0.49\textwidth}{!}{
    \begin{tabular}{l|cccc}
        \toprule
         \textbf{Language} & \textbf{Meaning} & \textbf{Written Variation} &
         \textbf{IPA} \\
         \midrule
         Javanese & what & apa / opo & \textipa{/OpO/} \\
         (Eastern-- & there is & ana / ono / onok & \textipa{/OnOP/} \\
         \textit{Ngoko}) & you & kon / koen & \textipa{/kOn/} \\
         \midrule
         Balinese & yes & inggih / nggih & \textipa{/PNgih/} \\
          (Alus--& I / me & tiang / tyang & \textipa{/tiaN/} \\
          \textit{Singgih})& <greeting> & swastyastu / swastiastu & \textipa{/swastiastu/} \\
         \midrule
         Sundanese & please / sorry & punten / punteun & \textipa{/punt@n/} \\
          (Badui--& red & beureum / berem & \textipa{/b@r1m/} \\
         \textit{Loma}) & salivating & ngacai / ngacay & \textipa{/NacaI/} \\
         \bottomrule
    \end{tabular}
    }
    \caption{Written form variations in several local languages, confirmed by native speakers.}
    \label{tab:written-form}
\end{table}

\subsection{Societal Challenges} \label{sec:societal-challenge}

Language evolves together with the speakers. A more widely used language may have a larger digital presence, which fosters a more written form of communication, while languages that are used only within small communities may emphasize the spoken form. Some languages are also declining, and speakers may prefer to use Indonesian
rather than their local language. In contrast, there are isolated residents that use the local language daily and are less proficient in Indonesian~\cite{nurjanah2018pengembangan,antara-NTT}. These variations give rise to different requirements, and there is no single solution for all.


Technology and education are not well-distributed within the nation. Internet penetration in Indonesia is 73.7\% in 2020 but is mainly concentrated on Java. Among the non-Internet users, 39\% explain that they do not understand the technology, while 15\% state that they do not have a device to access the Internet.\footnote{The Indonesian Internet Providers Association (APJII) survey: \url{https://apjii.or.id/survei2019x}} In some areas where the Internet is not seen as a basic need, imposing NLP technology on them may not necessarily be relevant. At the same time, general NLP development within the nation faces difficulties due to the lack of funding, especially in universities outside of Java. GPU servers are still scarce, even in top universities in the country.\footnote{For instance, we estimate the whole CS Faculty of the country's top university to have fewer than 10 GPUs.}


The dynamics of population movement in Indonesia also need to be taken into consideration. For example, urban communities transmigrate to remote areas for social purposes, such as teaching or becoming doctors for underdeveloped villages. Each of these situations might call for various new NLP technologies to be developed to facilitate better communication.

\section{Opportunities}

Based on the challenges for Indonesian NLP highlighted in the previous section, we formulate proposals for improving the state of Indonesian NLP research, as well as of other underrepresented languages. Our proposals cover several 
aspects including metadata documentation; 
potential research directions;
and engagement with communities.

\subsection{Better Documentation}

In line with studies promoting proper data documentation for NLP research~\cite{bender2018data, rogers-etal-2021-just-think, alyafeai2021masader, mcmillan2022documenting}, we recommend the following 
considerations.



\paragraph{Regional Dialect Metadata}
We have shown that a local language can have large variation depending on the region and the dialect. Therefore, we suggest adding regional dialect metadata to NLP datasets and models, not only for Indonesian but also for other languages. This is particularly important for languages with large dialectical differences. It also helps to clearly communicate NLP capabilities to stakeholders and end-users as it will help set an expectation of what dialects the systems can handle. Additionally, regional metadata can indirectly inform topics present in the data, especially for crawled data sources. 

\paragraph{Style and Register Metadata}

Similarly, we also suggest adding style and register metadata. This metadata can capture the politeness level of the text, not only for Indonesian but also in other languages.
In addition, this metadata can be used to document the formality level of the text, so it may be useful for research on modeling style or style transfer.

\subsection{
Potential Research Directions}


Among the most spoken local languages, a lot of research has been done on mainstream NLP tasks such as hate-speech detection, sentiment analysis, entity recognition, and machine translation. Some research has even been deployed in production by industry. 
Many of the languages, however, are not widely spoken and under-explored. Focusing on these languages, we suggest future research direction as follows.





\paragraph{Data-Efficient NLP}


Pretrained language models, which have taken the NLP world by storm, require an abundance of monolingual data. 
However, data collection has been a long-standing problem for low-resource languages. 
Therefore, we recommend more exploration into designing data-efficient approaches such as adaptation methods~\cite{artetxecross,aji2020neural, gururangan2020don,koto-etal-2021-indobertweet,kurniawan-etal-2021-ppt}, few-shot learning~\cite{winata2021language,madotto2021few, le2021many}, and learning from related languages \cite{Khanuja2021,khemchandani-etal-2021-exploiting}.
The goal of these methods is effective resource utilization, that is, to minimize the financial costs for computation and data collection as advocated by {\setcitestyle{citesep={,}}\citet{schwartz2020green,cahyawijaya2021greenformers}}, and \citet{nityasya2021costs}.\setcitestyle{citesep={;}}


\paragraph{Data Collection} Data collection efforts need to be commenced as soon as possible, despite all the challenges (\textsection \ref{sec:limited-res}). Here, we suggest collecting parallel data between Indonesian and each of the local languages for several reasons. 
First, a lot of Indonesians are bilingual~\cite{koto-koto-2020-towards}, that is, they speak both Indonesian and their local language, which facilitates data collection. Moreover, the fact that the local languages have some vocabulary overlap with Indonesian (see Table~\ref{tab:vocab_wiki} in the Appendix) might help facilitate building translation systems with relatively little parallel data~\cite{nguyen2017transfer}.
Finally, having such parallel data, we can build translation systems for synthetic data generation.
In line with this approach, the effectiveness of models trained on synthetic translated data can be explored.







\paragraph{Compute-Efficient NLP}

The costly GPU requirement for current NLP models hinders adoption by local research institutions and industries. Instead of focusing on building yet another massive model, we suggest focusing on developing lightweight and fast neural architectures, for example through distillation~\cite{kim2019research, sanh2019distilbert, jiao2020tinybert}, model factorization~\cite{winata2019effectiveness} or model pruning~\cite{voita2019analyzing}. 
We also recommend research on more efficient training mechanisms \cite{aji2017sparse, diskin2021distributed}. In addition, non-neural methods are still quite popular in Indonesia. Therefore, further research on the trade-off between the efficiency and quality of the models is also an interesting research direction.

\paragraph{Robustness to Code-mixing and Non-Standard Orthography}

Languages in Indonesia are prone to variations due to code-mixing and non-standard orthography, which occurs on the morpheme or even grapheme level. Models that are applied to Indonesian code-mixed data need to be able to learn morphologically faithful representations. Therefore, we recommend more explorations on methods derived from subword tokenization~\cite{gage1994bpe,kudo2018subword} and token-free models~\cite{gillick2016multilingual,tay2021charformer,xue2021byt5} to deal with this problem. This problem is also explored by~\citet{tan2021code} in an adversarial setting.

\paragraph{NLP Beyond Text}

For many Indonesian local languages that are rarely if ever written, speech is a more natural communication format. We thus recommend more attention on less text-focused research, such as spoken language understanding~\cite{chung2021splat,dmitriy2018slu}, speech recognition~\cite{besacier2014automatic,winata2020lrt,winata2020learning}, and multimodality~\cite{dai2020modality,dai-etal-2021-multimodal} in order to improve NLP for such languages.

\subsection{Engagement with Communities}

As discussed in \textsection \ref{sec:societal-challenge}, it is difficult to generalize a solution across local languages. We thus encourage the NLP community, such as the Indonesian Association of Computational Linguistics (INACL)\footnote{\url{https://inacl.id/inacl/}} to work more closely with native speakers and local communities.
Local communities who work on linguistics such as \textit{Polyglot Indonesia},\footnote{\url{http://polyglotindonesia.org}} \textit{Merajut Indonesia},\footnote{\url{https://merajutindonesia.id/}} and \textit{Masyarakat Linguistik Indonesia}\footnote{\url{https://www.mlindonesia.org/}} would be relevant collaborators to provide solutions and resources that support use cases benefiting the native speakers and communities of underrepresented languages. We advise the involvement of linguists, for example, to aid the language documentation process~\cite{anastasopoulos-etal-2020-endangered}. We also support open-science movements such as BigScience\footnote{\url{https://bigscience.huggingface.co/}} or ICLR CoSubmitting Summer\footnote{\url{https://blog.iclr.cc/2021/08/10/broadening-our-call-for-participation-to-iclr-2022/}} to help start collaborations and reduce the barrier to entry to NLP research.

\section{Conclusion}
In this paper, we highlight challenges in Indonesian NLP. Indonesia is one of the most populous countries and the second-most linguistically diverse country of the world, with over 700 local languages, yet Indonesian NLP is underrepresented and under-explored. 
Based on the observed challenges, we also present recommendations to improve the situation, not only for Indonesian but also for other underrepresented languages.

\section*{Acknowledgments}

We are grateful to Alexander Gutkin and Ling-Yen Liao for valuable feedback on an earlier draft of this paper. We also thank the anonymous reviewers for their helpful feedback. Lastly, we acknowledge the support of Kata.ai in this work.

\bibliography{anthology,custom}
\bibliographystyle{acl_natbib}

\clearpage

\appendix

\label{sec:appendix}

\section{Language Statistics}
\begin{figure}[!ht]
	\begin{subfigure}{0.98\linewidth}
		\centering
		\includegraphics[width=\linewidth]{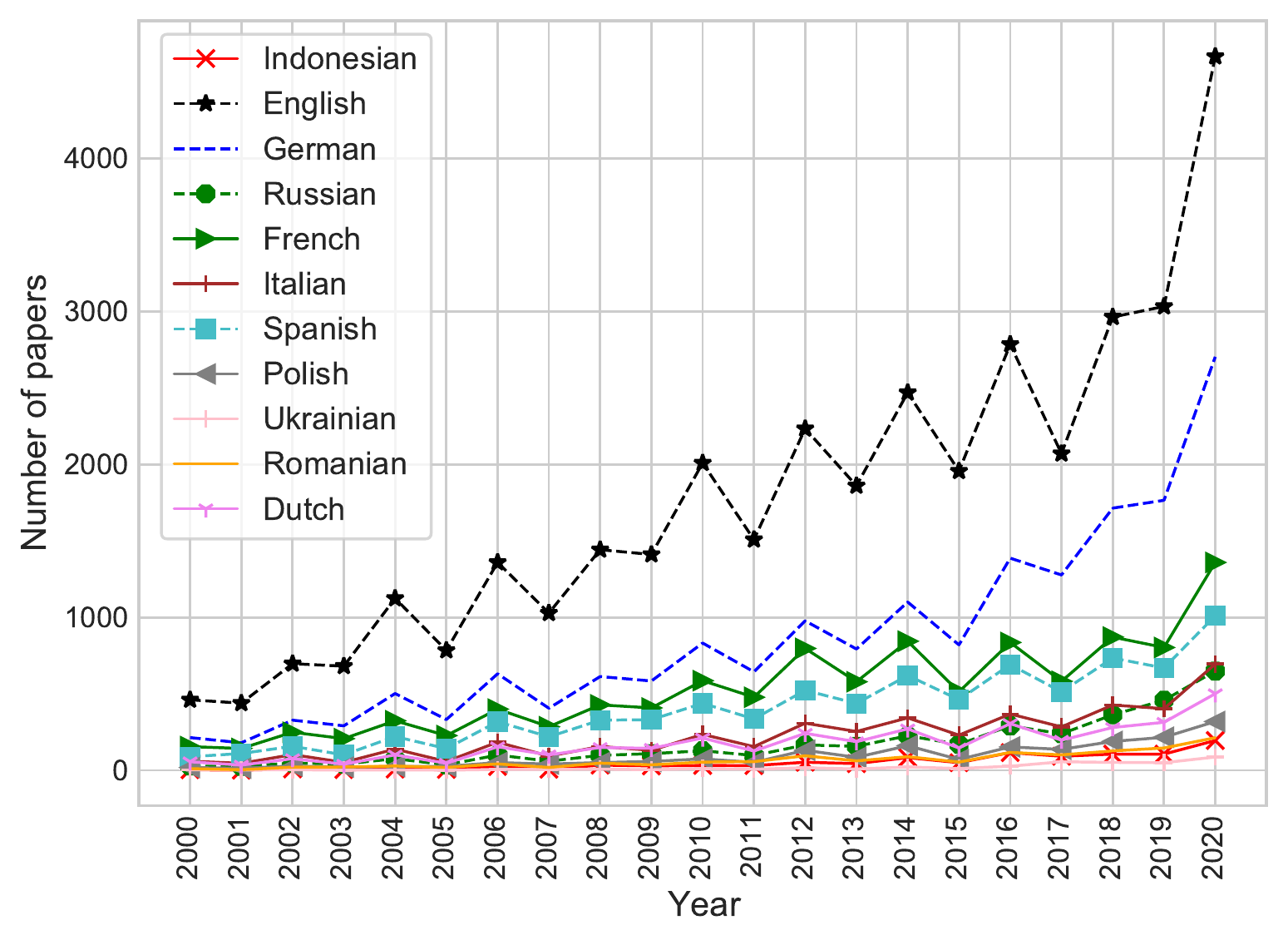} 
	\end{subfigure}
	\caption{Count of published papers per year on Indonesian and European languages from 2000 to 2020.}
	\label{fig:stats2}
	\vspace{-0.5cm}
\end{figure}

In \figref{stats2}, we contrast the count of publications related to
Indonesian languages compared to European languages. Although there are
many more Indonesian speakers than most European languages, the amount of published research relating to Indonesian is still comparatively lower than for European languages.

\section{Wikipedia Availability}
\label{appendix:wiki}

\begin{figure}[!th]
    \centering
    \includegraphics[width=\linewidth]{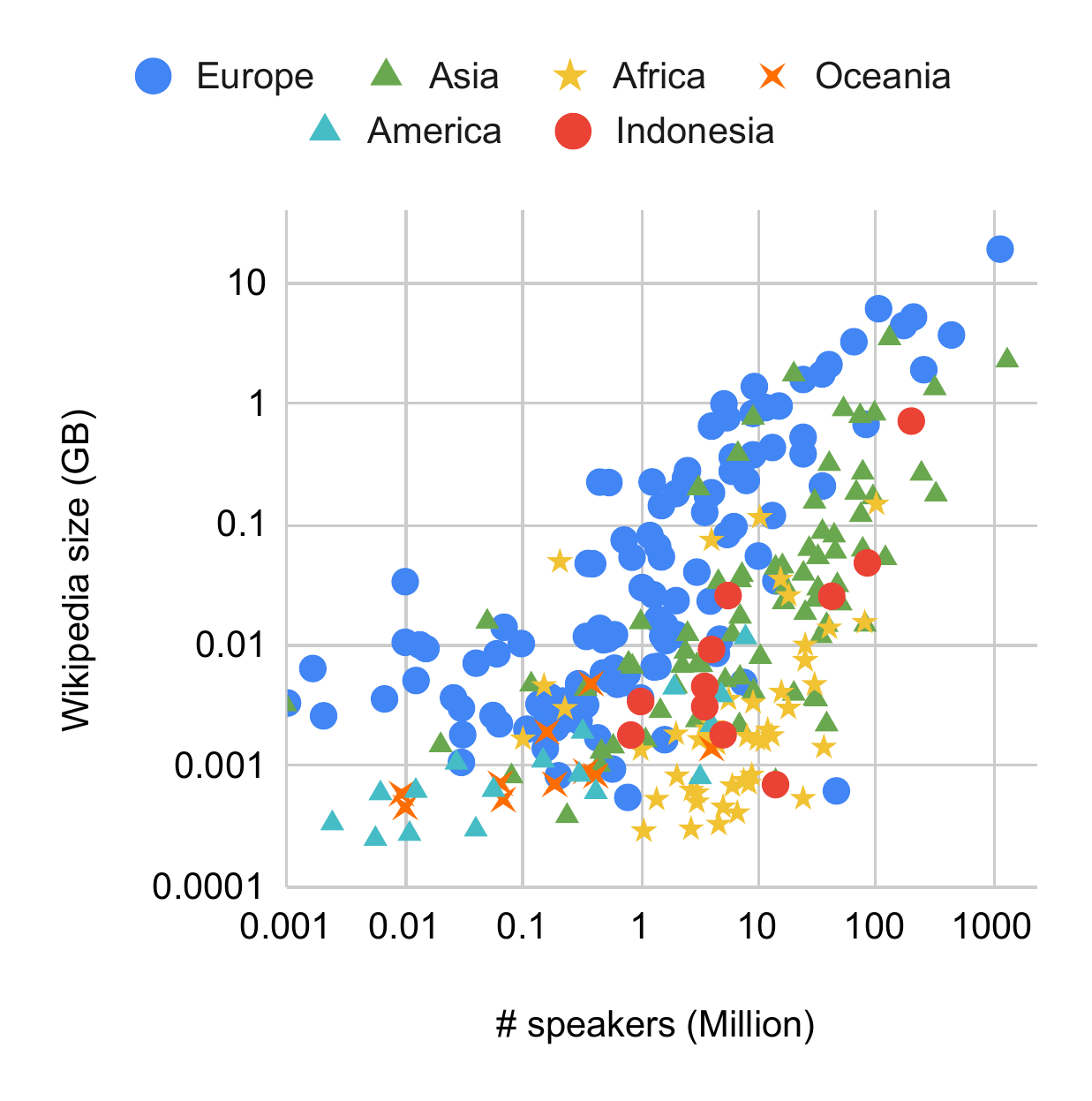}
    \caption{Wikipedia data size (in GB) compared to the number of speakers among Indonesia and different regions across the world.}
    \label{fig:wiki-all}
\end{figure}

In Figure~\ref{fig:wiki-all}, we compare Wikipedia size (in GB file size) compared to the number of speakers for various languages. We show that some African and indigenous American languages are even more under-resourced.

\section{Wikipedia Vocabulary Overlap}
In \tabref{vocab_wiki}, we present vocabulary statistics for Indonesian languages based on Wikipedia. Due to the noisy nature of Wikipedia, we use ``\textit{Kamus Besar Bahasa Indonesia}'' (KBBI) third edition,\footnote{\url{https://github.com/geovedi/indonesian-wordlist}} the official dictionary for the Indonesian language to filter out the top 1\% and top-100 most frequent words. As expected, the top 1\% words are less reliable, with only 59.3\% of vocabulary overlap between \texttt{id} and KBBI. In the top-100 words, there is a 96\% word overlap with KBBI, making this set more reliable. Previous work on Minangkabau by \citet{koto-koto-2020-towards} also showed that \texttt{id}-\texttt{min} words have a 55\% overlap in a manually curated bilingual dictionary, closer to the top-100 value for \texttt{min} in \tabref{vocab_wiki}.

\begin{table}[ht!]
    \centering
    \small
    \resizebox{0.9\linewidth}{!}{
        \begin{tabular}{lrrr}
        \toprule
            \multirow{4}{*}{\textbf{Lang}} & \multicolumn{3}{c}{\textbf{\# Vocab}} \\
            \cmidrule{2-4}
            & \multirow{2}{*}{All (k)} & Top 1\% $\cap$ & Top 100 $\cap$ \\
            & & KBBI (\%) & KBBI (\%) \\
        \midrule
            ind & 2023 & 59.3 & \textbf{96}\\
            jav & 435 & 46.8 & 43\\
            sun & 286 & 44.3 & 47 \\
            min & 252 & 30.3 & 41 \\
            bug & 23 & 35.7 & 27 \\ 
            map-bms & 14 & 76.7 & 79 \\
            gor & 12 & 40.5 & 49 \\
            ace & 12 & 37.6 & 46 \\
            ban & 10 & 43.3 & 46 \\
            bjn & 4 & 62.9 & 69 \\ 
            nia & 1 & 25.9 & 30 \\
            mad & 1 & 26.9 & 24 \\
        \bottomrule
        \end{tabular}
    }
    \caption{Vocabulary of Indonesian languages in Wikipedia, filtered with KBBI third edition.\footnotemark}
    \label{tab:vocab_wiki}
\end{table}
\footnotetext{KBBI is the official Indonesian dictionary.}

\begin{table}[!th]
    \centering
    \resizebox{0.99\linewidth}{!}{
    
    \begin{tabular}{@{}llll@{}}
        \toprule
        \textbf{English} & \textbf{Kedonganan} & \textbf{Jimbaran} & \textbf{Unggasan} \\
        \midrule
        I/me & Tyang & Tyang & Aku \\
        You & Béné & Béné & Éngko \\
        Umbrella & Pajéng & Pajéng & Pajong\\
        Hat & Capil & Topong & Cecapil, Tetopong \\
        How & Engken & Engken & Kengen \\
        Where & Dijé & Dijé & Di joho \\
        All & Konyangan & Onyé & Konyangan, onyang \\
        Swallow (vb) & Gélék,
ngélék & Gélék,
ngélék & Ngélokang\\
        Scratch (vb) & Gagas & Gagas & Gauk \\
        Cough (vb) & Kokoan & Dékah & Kohkohan\\
        Dawn & Plimunan & Plimunan & Sémongan\\
        Afternoon & Sanjé & Sanjé & Sanjano \\
        
        \bottomrule
    \end{tabular}
    }
    \caption{Lexical variation of Balinese across different villages in South Kuta district, Bali~\cite{maharani2018variasi}}
    \label{tab:bali-word}
    ~\\
    \resizebox{0.99\linewidth}{!}{
    
    \begin{tabular}{@{}l@{ ~ }l@{ ~ }l@{ ~ }l@{ ~ }l@{ ~ }l@{}}
        \toprule
        \multirow{2}{*}{\textbf{English}} & \multicolumn{1}{c}{\textbf{Pemenang}} & \multirow{2}{*}{\textbf{Jenggala}} & \multirow{2}{*}{\textbf{Genggelang}} & \multirow{2}{*}{\textbf{Kayangan}}  & \multirow{2}{*}{\textbf{Akar-Akar}}  \\
        & \multicolumn{1}{c}{\textbf{Timur}} \\
        \midrule
        Here & Ite & ite & ite & ite & tinI \\
        There & Ito & ito & ito & ito & tinO \\
        You & \textipa{diP} & sita  & \textipa{diP} & sita & \textipa{diP} \\
        Husband & \textipa{kur@nan} & sawa & sawa & sawa & sawa \\
        No & \textipa{deP} & \textipa{deP} & \textipa{deP} & \textipa{deP} & \textipa{soraP} \\
        Paddle & bose & bose & dayung & dayung & bose \\
        Spear & \textipa{t3r} & \textipa{cin@kan} & \textipa{t3r} & tombak & tombak \\
        Black & \textipa{bir@{\textrtailn}} & \textipa{bir@{\textrtailn}} & \textipa{bir@{\textrtailn}} & \textipa{bir@{\textrtailn}} & pisak\\
        Red & \textipa{b@n@{\textrtailn}} & \textipa{b@n@{\textrtailn}} & \textipa{b@n@{\textrtailn}} &  \textipa{b@n@{\textrtailn}} & \textipa{aba{\textrtailn}} \\
        White & \textipa{put3P} & \textipa{put3P}  & \textipa{put3P}  & \textipa{put3P} & \textipa{p@tak} \\
        Worm & \textipa{gumb@r} & \textipa{lo{\textrtailn}a}  & \textipa{gumb@r} & \textipa{gumb@r} & \textipa{gumb@r}  \\
        \bottomrule
    \end{tabular}
    }
    \caption{Lexical variation of Sasak across different villages in North Lombok district~\cite{sarwadi2019lexical}}
    \label{tab:sasak-word}
\end{table}

\section{Dialect Differences}
\label{appendix:dialect}

In this section, we present more examples of lexical variations in local Indonesian languages. \citet{maharani2018variasi} and \citet{sarwadi2019lexical} show lexical variations in Balinese and Sasak, respectively, where they asked locals to translate general/common words. Then, they compared the vocabulary across different locations (in this case, villages) to each other. Some of the examples can be seen in Tables~\ref{tab:bali-word} and \ref{tab:sasak-word}. Unfortunately, they did not provide quantitative results. \citet{pamolango2012geografi} conducted a similar experiment in the Banggai district of South Sulawesi across 31 observation points for the Saluan language. While~\citet{pamolango2012geografi} did not provide full examples, they reported up to 23.5\% lexical variation among 200 basic vocabulary items.

\section{Local Language Classification}

As shown in Table~\ref{tab:misclassification}, some Javanese texts are misidentified as Indonesian, English, and Malay. This is because Javanese and Indonesian (which is similar to Malay) share some words. We believe English misclassification is due to the data size bias. 

\begin{table}[!th]
    \centering
    \small
    \begin{tabular}{llllll}
    \toprule
        \textbf{Dialect/} &  & \multicolumn{4}{c}{\textbf{Class}} \\
        \textbf{Style} & Method & jav & idn & eng & mys \\
    \midrule
        Western- & Langid  & 0.241 & 0.103 & 0.172 & 0.069\\
        \textit{Ngoko} & FastText & 0.069 & 0.276 & 0.276 & 0.069\\
        & CLD3 & 0.759 & 0.000 & 0.000 & 0.034 \\
    \midrule
        Central- & Langid  & 0.345 & 0.138 & 0.069 & 0.069 \\
        \textit{Ngoko} & FastText & 0.379 & 0.310 & 0.069 & 0.069\\
        & CLD3 & 0.828 & 0.000 & 0.000 & 0.034\\
     \midrule
        Eastern- & Langid  & 0.276 & 0.103 & 0.069 & 0.138\\
        \textit{Ngoko} & FastText & 0.103 & 0.310 & 0.103 & 0.034\\
        & CLD3 & 0.552 & 0.103 & 0.000 & 0.000\\
     \midrule
        Eastern- & Langid  & 0.345 & 0.241 & 0.034 & 0.172 \\
        \textit{Krama} & FastText & 0.379 & 0.310 & 0.069 & 0.034 \\
        & CLD3 & 0.897 & 0.000 & 0.000 & 0.000\\
    \bottomrule
        
    \end{tabular}
    \caption{Language identification misclassification rate.}
    \label{tab:misclassification}
\end{table}

\label{app:nlp_resources2}
\begin{table*}[!ht]
    \centering
    \resizebox{0.99\textwidth}{!}{
    \small
    \begin{tabular}{llllrl}
    \toprule
    & \textbf{Paper} & \textbf{Language(s)} & \textbf{Domain/Source} & \textbf{Size} & \textbf{Link} \\ \midrule
    
    \multirow{2}{*}{\STAB{\rotatebox[origin=c]{90}{\scriptsize MONO}}} & \citet{wilie-etal-2020-indonlu} & ind & multi domain & ~3.6B & \href{https://github.com/indobenchmark/indonlu}{Indo4B} \vspace{0.15cm} \\
    
    & \citet{conneau-etal-2020-unsupervised} & ind & Web & 22.7B & \href{https://data.statmt.org/cc-100/}{CC--100 Indonesian} \\
    & & jav & Web & 24M & \href{https://data.statmt.org/cc-100/}{CC--100 Javanese} \\
    
    \midrule

    \multirow{8}{*}{\STAB{\rotatebox[origin=c]{90}{\scriptsize PARALLEL}}} & \citet{budiono2009resource} & ind $\leftrightarrow$ eng & News & 24k & \href{https://huggingface.co/datasets/id_panl_bppt}{PANL BPPT} \vspace{0.15cm}\\
    
     & \citet{larasati-2012-identic} & ind $\leftrightarrow$ eng & multi domain & 45k & \href{https://lindat.mff.cuni.cz/repository/xmlui/handle/11858/00-097C-0000-0005-BF85-F?show=full}{Identic} \vspace{0.15cm}\\
    
    & \citet{guntara-etal-2020-benchmarking} & ind $\leftrightarrow$ eng & Wikipedia & 93k & \href{https://github.com/gunnxx/indonesian-mt-data}{General En-Id} \\
    & & ind $\leftrightarrow$ eng & News & 42k & \href{https://github.com/gunnxx/indonesian-mt-data}{News En-Id} \\
    & & ind $\leftrightarrow$ eng & Religion & 590k & \href{https://github.com/gunnxx/indonesian-mt-data}{Religious En-Id} \vspace{0.15cm} \\ 
   
    & \citet{cahyawijaya-etal-2021-indonlg} & ind $\leftrightarrow$ eng & Religion & 31k & \\
    & & jav $\leftrightarrow$ eng & Religion & 16k & \href{https://github.com/indobenchmark/indonlg/tree/master/dataset/MT_JAVNRF_INZNTV}{Bible En-Jav} \\
    & & sun $\leftrightarrow$ eng & Religion & 16k & \href{https://github.com/indobenchmark/indonlg/tree/master/dataset/MT_SUNIBS_INZNTV}{Bible En-Sun}  \vspace{0.15cm} \\

    & \citet{koto-koto-2020-towards} & ind $\leftrightarrow$ min & Wikipedia & 16k & \href{https://github.com/fajri91/minangNLP/tree/master/translation/wiki_data}{MinangNLP MT} \vspace{0.15cm} \\
    
    & \citet{abidin2021effect} & ind $\leftrightarrow$ abl & Book & 3k & \href{https://drive.google.com/drive/folders/1oNpybrq5OJ_4Ne0HS5w9eHqnZlZASpmC}{Parallel: Indonesian -  Lampung Nyo}  \\
    
    \bottomrule
    
    \end{tabular}
    }
    \caption{List of monolingual and parallel corpora involving Indonesian and local languages. The size of the monolingual corpora is the number of words, while the size of the parallel corpora is the number of sentences.
    }
    \label{tab:corpora_stats}
\end{table*}

\section{Indonesian NLP Resources}
\label{appendix:resource}

In Table~\ref{tab:corpora_stats}, we present the list of existing monolingual and parallel corpora in Indonesian and local languages. Table~\ref{tab:labeled_dataset_ina} shows the list of publicly available NLP datasets for Indonesian. Table~\ref{tab:tools_and_resources_ina} shows the list of NLP tools and resources for Indonesian. Table~\ref{tab:labeled_dataset_local} shows the list of publicly available NLP datasets for local languages spoken in Indonesia.

Although the volume of data in local languages is much smaller than that for Indonesian, these resource collections are arguably beneficial for constructing resources in other local languages. This is because: (1) Indonesian can be used as a pivot language with regard to local languages, due to the large vocabulary overlap (see \tabref{vocab_wiki}); and 2) most Indonesians are bilingual, speaking both Indonesian and their local language \cite{koto-koto-2020-towards}.





\label{app:nlp_resources}
\begin{table*}[!ht]
    \centering
    \resizebox{0.99\textwidth}{!}{
    \small
    \begin{tabular}{lllrlp{500pt}}
    \toprule
    \textbf{Work by} & \textbf{Task} & \textbf{Domain/Source} & \textbf{Size} & \textbf{Dataset link}  \\ \midrule
	
	\citet{pimentel-ryskina-etal-2021-sigmorphon} & Morphology Analysis & Dictionary, Wikipedia & 27k & \href{https://github.com/unimorph/ind}{unimorph id} \\
	
	\citet{dinakaramani_pos_tag} & POS Tagging & News & 10k & 
	\href{https://github.com/famrashel/idn-tagged-corpus}{POS bahasa.cs.ui.ac.id} \\
	\citet{hoesen2018investigating} & POS Tagging & News & 8k & \href{https://github.com/indobenchmark/indonlu}{IndoNLU/POSP}
	\\
	 \citet{hoesen2018investigating} & Named Entity Recognition &  News & 8k & 
	 \href{https://github.com/indobenchmark/indonlu}{IndoNLU/NERP}  \\

    NERGrit & Named Entity Recognition & News & 23k & 
     \href{https://github.com/grit-id/nergrit-corpus}{nergrit-corpus} \\

	\citet{fachri} & Named Entity Recognition & News & 2k & 
	\href{https://github.com/yusufsyaifudin/indonesia-ner}{Indonesian NER} \\

	\citet{mdee} & Named Entity Recognition & Wikipedia & 48k & \href{https://github.com/ialfina/ner-dataset-modified-dee}{Singgalang modified-dee} \\

    \citet{mahendra-etal-2018-cross} & Word Sense Disambiguation & multi domain & 2k & \href{https://github.com/rmahendra/Indonesian-WSD}{Indonesian WSD} \\

    \citet{moeljadi_jati:2017} & Constituency Parsing & Dictionary & 1.2k & \href{https://github.com/davidmoeljadi/INDRA/tree/master/tsdb/gold/jati_edited}{JATI} \\
    
    \citet{moeljadi2019building} & Constituency Parsing & Chat & 0.7k & \href{https://github.com/davidmoeljadi/INDRA/tree/master/tsdb/gold/Cendana}{Cendana} \\
    
    \citet{kethu} & Constituency Parsing & News & 1k & \href{https://github.com/ialfina/kethu}{kethu} \\

    \citet{mcdonald-etal-2013-universal} & Dependency Parsing & News, Blog & 5k & \href{https://github.com/UniversalDependencies/UD_Indonesian-GSD}{UD\_Indonesian-GSD} \\

    \citet{zeman-etal-2018-conll} & Dependency Parsing & News, Wikipedia & 1k & \href{https://github.com/UniversalDependencies/UD_Indonesian-PUD}{UD\_Indonesian-PUD} \\

    \citet{artari-etal-2021-multi} & Coreference Resolution & Wikipedia & 0.2k & \href{https://github.com/valentinakania/indocoref}{IndoCoref} \\

    \citet{purwarianti2007qa} & Question Answering & News & 3k & \href{https://github.com/indobenchmark/indonlu}{IndoNLU/FacQA}
    \\ 

    \citet{clark-etal-2020-tydi} & Question Answering & Wikipedia & 18k & \href{https://github.com/google-research-datasets/tydiqa}{tydiqa/id} \\

    \citet{kurniawan2018indosum} & Summarization & News & 20k & \href{https://github.com/kata-ai/indosum}{IndoSum} \\

    \citet{koto-etal-2020-liputan6} & Summarization & News & 215k & \href{https://github.com/fajri91/sum_liputan6}{Liputan6} \\

    \citet{mahfuzh2019improving} & Keyphrases Extraction & Twitter & 1k & \href{https://github.com/indobenchmark/indonlu}{IndoNLU/KEPS} \\

    \citet{ken2018entailment} & Natural Language Inference & Wikipedia & 0.5k & \href{https://github.com/indobenchmark/indonlu}{IndoNLU/WreTe} \\

    \citet{mahendra-etal-2021-indonli} & Natural Language Inference & Wikipedia, News & 18k & \href{https://github.com/ir-nlp-csui/indonli}{IndoNLI} \\ 

    \citet{purwarianti2019improving} & Sentiment Analysis & Review & 13k  & \href{https://github.com/indobenchmark/indonlu}{IndoNLU/SmSA} \\
     \citet{ilmania2018absa} & Sentiment Analysis & Review & 1k  & \href{https://github.com/indobenchmark/indonlu}{IndoNLU/CASA} \\
    \citet{azhar2019multi} & Sentiment Analysis & Review & 3k &  \href{https://github.com/indobenchmark/indonlu}{IndoNLU/HoASA} \\

    \citet{koto-etal-2020-indolem} & Sentiment Analysis & Twitter, Review & 5k & \href{https://github.com/indolem/indolem}{IndoLEM/sentiment} \\

    \citet{saputri2018emotion} & Emotion Classification & Twitter & 4k & \href{https://github.com/meisaputri21/Indonesian-Twitter-Emotion-Dataset}{Indonesian emotion}
    \\ 
    \citet{jannati2018stance} & Stance Detection & Blog & 0.3k & \href{https://github.com/reneje/id_stance_dataset_article-Stance-Classification-Towards-Political-Figures-on-Blog-Writing}{Indonesian stance} 
    \\ 
    \citet{alfina2017hate} & Hate Speech Detection & Twitter & 0.5k & \href{https://github.com/ialfina/id-hatespeech-detection}{id hatespeech} \\
    \citet{IBROHIM2018} & Hate Speech Detection & Twitter & 2k & \href{https://github.com/okkyibrohim/id-abusive-language-detection}{id abusive} \\
    \citet{ibrohim-budi-2019-multi} & Hate Speech Detection & Twitter & 13k & \href{https://github.com/okkyibrohim/id-multi-label-hate-speech-and-abusive-language-detection}{id multilabel HS} \\
    \citet{WILLIAM2020clickbait} & Clickbait Detection & News & 15k & \href{https://data.mendeley.com/datasets/k42j7x2kpn/1}{Indonesian clickbait} 
    \\ 
    \citet{wibowo2020semi} & Style Transfer & Twitter & 2k & \href{https://github.com/haryoa/stif-indonesia}{STIF-Indonesia}\\
	\bottomrule
    \end{tabular}
    }
    \caption{List of publicly available NLP datasets for the Indonesian language. The size of the dataset is defined by the number of sentences in most cases, except for morphology analysis (number of words), coreference resolution, summarization, and stance detection (number of articles).
    }
    \label{tab:labeled_dataset_ina}
\end{table*}

\begin{table*}[!ht]
    \centering
    \resizebox{0.99\textwidth}{!}{
    \small
    \begin{tabular}{llp{280pt}}
    \toprule
    \textbf{Resource Name} & \textbf{Reference} & \textbf{Description} \\ \midrule
	
	\href{https://github.com/sastrawi/sastrawi}{Sastrawi} & -- & Indonesian stemming implementation based on~\citet{adriani2007stemming}'s algorithm \\
    
    \href{https://github.com/panggi/pujangga}{Pujangga} & -- & The interface for InaNLP ~\cite{purwarianti2016inanlp}, an Indonesian NLP toolkit \\
    
    \href{https://github.com/davidmoeljadi/INDRA}{INDRA} & \citet{Moeljadi_dkk15} & Indonesian resource grammar  \\
    
    \href{https://septinalarasati.com/morphind/}{MorphInd} & \citet{larasati2011indonesian} & Morphology tool  \\
    
    \href{https://github.com/matbahasa/MALINDO_Morph}{MALINDO Morph} & \citet{nomoto2018malindo} & Morphological dictionary and analyzer \\
    
    \href{https://github.com/bahasa-csui/aksara}{Aksara} & \citet{hanifmuti2020aksara} & Indonesian morphological analyzer based on the UD v2 annotation guidelines \\
    
    \href{https://bahasa.cs.ui.ac.id/iwn/}{WordNet IWN} & \citet{putra2008building} & WordNet (UI), consisting of 1,203 synsets and 1,659 words in Indonesian \\
    
    \href{http://wn-msa.sourceforge.net/}{WordNet Bahasa} & \citet{Bond2014} & WordNet (NTU), consisting of 49,668 synsets and 64,131 unique words in Malay and Indonesian \\
    
    \href{https://github.com/fajri91/InSet}{Inset} & \citet{koto2017inset} & Indonesian sentiment lexicon 
     of 3,609 positive and 6,609 negative 
    words 
    \\
	\href{https://github.com/masdevid/ID-OpinionWords}{masDevid opinion word} & \citet{wahid2016peringkasan} & List of Indonesian opinion words (1,182 positive and 2,402 negative), translated from ~\citet{liu2005opinion}
    \\
	
	\href{https://github.com/nasalsabila/kamus-alay}{Kamus Alay} & \citet{salsabila2018lexicon} &  Indonesian formal to informal lexicon \\
	\href{https://github.com/haryoa/indo-collex}{IndoCollex} & \citet{wibowo-etal-2021-indocollex} & Indonesian formal to informal lexicon, categorized by the transformation type \\
	 \bottomrule
    \end{tabular}
    }
    \caption{List of Indonesian NLP tools and resources. 
    }
    \label{tab:tools_and_resources_ina}
\end{table*}

\begin{table*}[!ht]
    \centering
    \resizebox{0.99\textwidth}{!}{
    \small
    \begin{tabular}{llllrlp{500pt}}
    \toprule
    \textbf{Work by} & \textbf{Task} & \textbf{Language(s)} & \textbf{Domain/Source} & \textbf{Size} & \textbf{Dataset link}  \\ \midrule
	
	\citet{koto-koto-2020-towards} & Sentiment analysis & min & Twitter, review & 5000 & \href{https://github.com/fajri91/minangNLP}{Minang NLP sentiment} \\
	\citet{putra2020sundanese} & Emotion classification & sun & Twitter & 2518 & \href{https://github.com/virgantara/sundanese-twitter-dataset}{Sundanese emotion}  \\
	
	\citet{Putri_etal_2021_abusive} & Hate speech detection & jav & Twitter & 3478 & \href{https://github.com/Shofianina/local-indonesian-abusive-hate-speech-dataset}{Javanese HS}   \\
	\citet{Putri_etal_2021_abusive} & Hate speech detection & sun & Twitter & 2209 & \href{https://github.com/Shofianina/local-indonesian-abusive-hate-speech-dataset}{Sundanese HS}\\
	Javanese NLP CSUI & Dependency parsing & jav & Wikipedia & 125 & \href{https://github.com/UniversalDependencies/UD_Javanese-CSUI}{UD Javanese CSUI} \\
	\citet{wongso2021causal} & Sentiment analysis & jav & Review & 100k & \href{https://huggingface.co/datasets/w11wo/imdb-javanese}{Javanese translated IMDB} 
	\\
	\bottomrule
    \end{tabular}
    }
    \caption{List of publicly available NLP datasets for local languages spoken in Indonesia. The size of the dataset is defined by the number of sentences.
    }
    \label{tab:labeled_dataset_local}
\end{table*}

\clearpage

\end{document}